\documentclass[11pt]{asaproc}

\usepackage{graphicx}
\usepackage{amsmath}
\usepackage{amssymb}
\usepackage{amsfonts}
\usepackage{bbm}
\usepackage{amsthm}
\usepackage[ruled,vlined]{algorithm2e}
\usepackage{stmaryrd}
\usepackage{appendix}

\usepackage[
backend=biber,
style=alphabetic,
sorting=ynt
]{biblatex}
\newtheorem{theorem}{Theorem}

\newcommand{\R}{{\mathbb{R}}}
\newcommand{\E}{{\mathbb{E}}}

\title{ Deep Upper Confidence Bound Algorithm for Contextual Bandit Ranking of Information Selection }

\author{Michael Rawson\thanks{
Corresponding Author; Department of Mathematics, University of Maryland,
College Park, MD, USA,
rawson@math.umd.edu 
} \and  Jade Freeman\thanks{
DEVCOM Army Research Laboratory, Adelphi, MD, USA,
jade.l.freeman2.civ@army.mil
} }
\begin{document}

\maketitle 

\begin{abstract} 
Contextual multi-armed bandits (CMAB) have been widely used for learning to filter and prioritize information according to a user’s interest. In this work, we analyze top-$K$ ranking under the CMAB framework where the top-$K$ arms are chosen iteratively to maximize a reward. The context, which represents a set of observable factors related to the user, is used to increase prediction accuracy compared to a standard multi-armed bandit. Contextual bandit methods have mostly been studied under strict linearity assumptions, but we drop that assumption and learn non-linear stochastic reward functions with deep neural networks. We introduce a novel algorithm called the Deep Upper Confidence Bound (UCB) algorithm. Deep UCB balances exploration and exploitation with a separate neural network to model the learning convergence. We compare the performance of many bandit algorithms varying K over real-world data sets with high-dimensional data and non-linear reward functions. Empirical results show that the performance of Deep UCB often outperforms though it is sensitive to the problem and reward setup. Additionally, we prove theoretical regret bounds on Deep UCB giving convergence to optimality for the weak class of CMAB problems. 
\begin{keywords}
Multi-Armed Bandit, Contextual Bandit, Regret, Deep Learning, Neural Network, Ranking, Recommendation, Meta-Learning
\end{keywords}
\end{abstract}

\section{Introduction}

In the current era of perpetual influx of information, decision makers and analysts are faced with challenges under limited time and resources. To lighten the load, a selection policy is used to filter out the information that is of little use or interest to the user. The Multi-Armed Bandit (MAB) \cite{Gittins1979BanditPA}, \cite{Bather1990MultiArmedBA} framework (see section 2) can be employed to dynamically match an object to the user's interest based on previously seen behavior. Further, contextual bandit allows the learning process to incorporate features and context regarding the user, the objects, and the decision-making environment. In contextual bandit, assuming a linear relationship between the context and the reward is often sufficient and simple and straightforward in implementation. However, it can result in unsatisfactory performance when linearity assumption is invalid. Deep neural networks (DNN) have been successful in approximating non-linear value functions, are proven to be a universal function approximator, and can learn associations between complex states to estimate expectations \cite{Riquelme2018DeepBB}. Further, DNNs allow generalization of the states without domain specific knowledge and learn rich domain representations from raw, high-dimensional inputs \cite{Mnih2015HumanlevelCT}, \cite{Silver2016MasteringTG}, \cite{Zahavy2019DeepNL}. Deep bandit learning has been studied in a Bayesian framework, which utilizes uncertainty on the model in exploration \cite{Chen2013CombinatorialMB}, \cite{Guo2020DeepBB}. 

In this work, we introduce a novel algorithm called Deep Upper Confidence Bound (UCB). Deep UCB can learn the optimal solution to the top-$K$ ranking problem. Deep UCB balances exploration and exploitation with a separate neural network to model the learning convergence.  This paper is organized as follow: Section 2 discusses previous work, Section 3 formalizes the problem, Section 4 describes our algorithm, Section 5 shows performance comparisons on real world experiments, Section 6 gives a regret bound analysis, and Section 7 is the conclusion.

\section{Previous Work}

The Multi-Armed Bandit (MAB) problem is a iterative decision problem where a decision maker is presented with multiple arms and must choose the optimal arm(s) repetitively  \cite{Gittins1979BanditPA}, \cite{Bather1990MultiArmedBA}. In applications, the ``arms" represent a set of objects with each having an unknown reward upon selection \cite{Qin2014ContextualCB}. MAB framework can be applied to the reinforcement learning problem of dynamically matching an object to the user's interest from the previously seen behavior. Contextual bandits incorporate some type of context information to improve matching ability. A reward function is used to measure the quality or relevance of the selected object based on the observed user feedback. The goal of the learning algorithm is to maximize the rewards or equivalently to minimize regrets over time.

Bandit algorithms employ a selection strategy that balances choices between exploration of the untested arms and exploitation of previously successful or high reward arms. One of the most popular MAB strategies \cite{Langford2007TheEA} is the $\epsilon$-greedy method that conducts exploration uniformly over arms with $\epsilon > 0$ probability and chooses the best arm on average with $1-\epsilon$ probability. One drawback of $\epsilon-greedy$ is that it does not converge to optimal strategy/arm. By making the $\epsilon$ decay time, exploration diminishes as the learner becomes confident. This will converge to the optimal arm with high probability. 

Another well-known MAB strategy is Thompson sampling \cite{Thompson1933ONTL}, \cite{Agrawal2012AnalysisOT}. Instead of averaging the rewards of a specific arm, an entire distribution is inferred. In particular, the previous rewards are used to estimate the parameters of that arm's reward distribution. For each arm distribution, a sample is drawn. The arm with the highest sample is chosen. This strategy converges but is limited if the distribution type is unknown. 

When the reward of each arm depends on some varying contextual information, it is called a contextual bandit (see section 3). The context is modeled as a vector that is provided to the algorithm at each iteration or time step. Contextual Bandit algorithms include Linear Regression, LinUCB, Deep epsilon greedy, Neural-linear, and Deep Thompson. The linear regression algorithm performs linear regression to map each context vector to its expected reward for each arm. The LinUCB \cite{li_contextual-bandit_2010} algorithm models a linear function from the context vector to the expected reward plus a term representing the uncertainty for each arm. As in the linear regression, this method assumes linear reward functions and lacks exploration for nonlinear models. For example, a 0 context vector will always predict expectation 0 even though the expectation could be any real number. Furthermore, the computation time can be prohibitively high in the range of cubic complexity due to matrix inversion. Deep $\epsilon$-greedy implements the $\epsilon$-greedy method using a neural network to approximate the expected reward function. Neural-linear method combines neural network mapping of each context vector to some latent vector and then linear regression mapping the latent vector to the predicted expected reward. This is equivalent to using a neural network when optimizing mean squared error. In practice, neural-linear method converges faster than a neural network because training a standard neural network does not reach optimality. Since Neural-linear optimizes very quickly, there is little exploration comparatively.

\section{Contextual Bandits}

The contextual bandit is the multi-armed bandits (MAB) problem where given a measurable context vector in $\mathcal{C}$ that changes over time, the one item of a static list must be chosen at each time. The list of items are called ``arms" in the MAB framework. The selection of the arms has cost function or reward function and the goal of the bandit method is to choose the optimal policy that maximizes the reward function $R$ given the context vector on the item, $\mathcal{C}$. To formalize this, the contextual MAB method finds the best context mapping policy
$\pi : \mathcal{H} \times \mathbb{T} \times \R^m \rightarrow A$ 
where time
$\mathbb{T} = [1,T] \cap \mathbb{N}$ and
$A = \{a_1,...,a_n\}$ is the set of the arms that can be chosen and $H \in \mathcal{H}$ is the history of realizations of random variables $R$ and $C$. We will allow the reward function $R$ and context vector $\mathcal{C}$ to be stochastic to account for noise in the system. We assume the realizations, or samples, of $R$ and $\mathcal{C}$ are i.i.d respectively. Then random variables $R : A \times \R^m \times \Omega_1 \rightarrow \R$ and $\mathcal{C} : \Omega_2 \rightarrow \R^m$ with sample space $\Omega_1$ and $\Omega_2$ respectively. The optimal policy then needs to maximize the \emph{expectation} of the rewards over time. We'll take the expectation over $R$ and $\mathcal{C}$ over each time step. The optimal policy
$$ \pi^* = \arg\max_{\pi}  \E_{\omega_1\in\Omega_1^T,\omega_2\in\Omega_2^T}
\sum_{t=1}^T  \ R(\pi(H^{(t)}_{\omega_{1},\omega_{2}},t,\mathcal{C}(\omega_{2,t})), \mathcal{C}(\omega_{2,t}), \omega_{1,t})$$ 
where $\Omega^T$ is the $T$ product space and $\omega_i$ is a vector with index $t$ given by $\omega_{i,t}$ and the measure is the product measure. We denote the history up to time $t$ by $H^{(t)}_{\omega_{1},\omega_{2}}$ which drops $\omega_{1,j},\omega_{2,j}$ for $j \ge t$. The optimal solution, for any $H \in \mathcal{H}, t\in \mathbb{T}, \omega_2 \in \Omega_2$, has 
$ \pi^*(H,t,\mathcal{C}(\omega_2)) = \arg\max_{a\in A} \E_{\omega_1\in\Omega_1} R(a,\mathcal{C}(\omega_2),\omega_1) $. Typically, the distributions and expectations are not known, so methods will use the history to estimate an optimal solution. 

\subsection{Top-$K$ Ranking}

In this section, we describe using contextual bandits for top-$K$ ranking. The top-$K$ ranking is based on optimally sorting lists iteratively. For top-$K$ ranking, the highest k items or arms must be chosen in each iteration. There is no a priori information about how to rank or choose the top k arms. The ranking function must be learned by receiving rewards for each top-$K$ arm chosen. When $k=1$, this is the regular multi-armed bandit (MAB). When $k>1$, we use MAB methods to rank the arms and the top k arms are chosen. 

When the reward is a function of context vectors that change over time steps or iterations, we use contextual multi-armed bandit (CMAB) methods. The CMAB is just the MAB problem additionally given the context vector at each iteration where the reward may depend on the context vector. So the CMAB method needs to approximate a reward function that maps the context vector and arm pair to a real number. Two common ways to do this are with linear models (e.g., linear regression) or neural networks. In CMAB, assuming a linear relationship between the context and the reward is simple and straightforward in implementation. However, it can result in unsatisfactory performance when the linearity assumption is invalid. The deep neural networks (DNN) can efficiently approximate non-linear value functions \cite{Riquelme2018DeepBB}. Further, DNNs can generalize the contextual states without domain specific knowledge and learn latent representations from raw, high-dimensional data \cite{Mnih2015HumanlevelCT}, \cite{Silver2016MasteringTG}, \cite{Zahavy2019DeepNL}. Deep bandit learning has been studied in a Bayesian framework, which utilizes uncertainty on the model in exploration \cite{Chen2013CombinatorialMB}, \cite{Guo2020DeepBB}. In the following section, we describe a novel algorithm called the Deep Upper Confidence Bound (UCB) algorithm. Deep UCB balances exploration and exploitation with neural network models. 

\section{Deep Upper Confidence Bound for CMAB}

Deep Upper Confidence Bound (Deep UCB) method for high dimensional, nonlinear CMAB problems is motivated by Upper Confidence Bound (UCB) method \cite{auer_finite-time_2002} for MAB where the arm with the highest expected reward plus uncertainty is chosen. Deep UCB uses two neural networks. One neural network predicts the expected reward and the second neural network predicts uncertainty. The arms with the highest expected rewards plus high uncertainty are chosen. Unlike the other methods, nonlinear reward functions are accurately modeled and exploration is guided to maximize certainty. Choosing uncertain arms decreases the uncertainty. As uncertainty of all arms approaches zero, this method converges to choosing the arm with the highest expected reward. 

We model the expectation of reward functions, $\mu_c , c \in \mathcal{C}$, from a neural network where the input is the context vector $c \in \mathcal{C}$ and the output is a vector approximating the reward, $NN : \mathcal{C} \rightarrow \R$. Further, we estimate the uncertainty of the expected reward, $\sigma_c$, given a context vector $c \in \mathcal{C}$, represented as the standard error of the expectation. This procedure is done for each arm. Then, for each arm, the deep upper confidence bound (DUCB) estimate is defined, given $c \in \mathcal{C}$, as
$$ DUCB(c) = \hat\mu_c + \hat\sigma_c/\sqrt{n} $$
 by the arm's $n$ realizations, empiric expectation $\hat\mu_c$, and empiric standard deviation $\hat\sigma_c$. Here, $\mu$ and $\sigma$ are unknown functions of the context vector; $\mu,\sigma : \mathcal{C} \rightarrow \R$. While we estimate $\hat\mu_c$ approximately with $NN_1 = \hat\mu$, we approximate $\sigma$ with another neural network, $NN_2 : \mathcal{C} \rightarrow \R$ and $NN_2 = \hat\sigma^2$, and thus, 
$$ DUCB(c) = NN_1(c) + \sqrt{NN_2(c)/n} $$
where $n$ is number of realizations of the arm and $c$ is the context vector. As implemented, the neural networks have multiple outputs and will compute the prediction for each arm simultaneously and then calculation is as above but with vectors. 

When training these neural networks, the mean squared error (MSE) of $NN_1$ is minimized and the $L^1$ error of $NN_2$ is minimized. Let $R(c)$ be a realization of the reward of a context vector $c$ for an arm and $S \subset \mathcal{C}$ be some subset of the context vectors. Then, for a sufficiently parameterized NN,
$$ NN_1 = \arg\min_{NN_1} \sum_{c \in S} (NN_1(c) - R(c))^2 $$ 
is the maximum likelihood esimator for each $c$ and converges when the reward function $R(c) \sim $ Gaussian. Likewise,
$$ NN_2 = \arg\min_{NN_2} \sum_{c \in S} | NN_2(c) - (NN_1(c)-R(c) )^2 |. $$ 
For each $c$, with $T$ independent realizations of $R$ and $t=1,...,T$, 
$$  \sum_{ 1\le t\le T } |NN_2(c) - (NN_1(c)-R_t(c))^2| $$
$$ =  \sum_{ t\in T_1 } NN_2(c) - (NN_1(c)-R_t(c))^2 
-\sum_{ t\in T_2 } NN_2(c) - (NN_1(c)-R_t(c))^2 $$
where $T_1 = \{1 \le t \le T : NN_2(c) - (NN_1(c)-R_t(c))^2 \ge 0 \} $ and $T_2 = [1,T] \backslash T_1$. As $NN_1(c)\rightarrow \mu_c$,
$$ \sum_{ t \in T_1 } NN_2(c) - (NN_1(c)-R_t(c))^2
\rightarrow  T_1 \cdot NN_2(c) - \sum_{ t \in T_1 } (\mu_c - R_t(c))^2 $$ 
Likewise for $T_2$. By minimizing $ \sum_{ T } |NN_2(c) - (NN_1(c)-R(c))^2| $, then $NN_2(c) \rightarrow \sigma_c^2 $ as $NN_1(c)\rightarrow \mu_c$ and $T \rightarrow \infty$.
Therefore, DUCB converges to the upper confidence bound given sufficiently parameterized neural networks and enough training samples. 

We'll call this method Deep UCB2, given in Algorithm \ref{ducb2}. Note that by assuming the arms are similar stochastic functions (but still unknown), we've simplified Algorithm \ref{ducb2}.  We call a variation of this method Deep UCB1, given in Algorithm \ref{ducb1}, which has an accompanying regret bound analysis in section 6.

\begin{algorithm}[h]
\SetAlgoLined
$ C \in \mathbb{R}^{T\cdot K \cdot m}$ where $C_{t,j}$ is the $j^{th}$ context vector in $\mathbb{R}^m$ at time step $t$  \\ 
$ A = \{arm_1,...,arm_N\}$  \\
$ NN_{1,Z_t}, NN_{2,Z_t} = $ Neural Network $:\mathbb{R}^m \rightarrow \mathbb{R} $ with $Z_t$ neurons \\
$ Reward : \mathbb{N}_{[1,K]} \rightarrow \mathbb{R}$ \\
$ \epsilon_i = \mu_{i,max}-\mu_{i,min} + \epsilon $ \\
 \For{t = 1 ... T}{
  \uIf{$t \le JN$}{ 
    \# Initial exploration phase \\
    \For{j = 1 ... K}{
      \# Choose arms \\
      $D_{t,j} =  ((\lfloor \frac{t-1}{J} \rfloor + j - 1) \mod N)+1$ \\
    }
  }
  \Else{
      \For{j = 1 ... K}{
       \# $NN(\cdot)$ gives a prediction using NN \\
       $ \hat{\mathbb{E}}_R(arm_j) = \frac{1}{W_t} \sum_{i=1}^{W_t} NN_{R,Z_t}^{(i)}(C_{t,j}) $ \quad \# Predict reward \\
       $ \hat{\mathbb{E}}_V(arm_j) = \frac{1}{W_t} \sum_{i=1}^{W_t} NN_{V,Z_t}^{(i)}(C_{t,j}) $ \quad \# Predict certainty \\
      }
      \For{j = 1 ... K}{
        \# Choose arms \\
        $D_{t,j} =  \arg \max_{arm \in A - D_{t,:} } \ \hat{\mathbb{E}}_R(arm)
        + \sqrt{\frac{2\hat{\mathbb{E}}_V(arm) + 2 \ln t }{ \sqrt{|\{d \in D_{1:t-1,:} \ s.t. \ d=arm\}|} } } + \epsilon_{arm} $ \\
      }
      \For{j = 1 ... K}{
        $R_{t,j} = Reward(D_{t,j})$ \quad \quad \quad \ \# Record rewards \\
        $V_{t,j} = (R_{t,j}-\hat{\mathbb{E}}_R(arm_j))^2 $ \quad \# Record accuracy \\
      }
      \# Train the networks \\
      \For{i = 1 ... $W_t$}{
        $ J_i = \llbracket (i-1)\frac{t}{W_t}+1 , i\frac{t}{W_t} \rrbracket $ \\ 
        TrainNNet($NN_{R,Z_t}^{(i)},\ in=C_{J_i,D_{t,1:K}},\ out=R_{J_i,1:K},$ loss=MSE)  \\
        TrainNNet($NN_{V,Z_t}^{(i)},\ in=C_{J_i,D_{t,1:K}},\ out=V_{J_i,1:K},$ loss=MSE) \\
      }
    }
  }
 \caption{Deep UCB1}
 \label{ducb1}
\end{algorithm}

\begin{algorithm}[h]
\SetAlgoLined
$ C \in \mathbb{R}^{T\cdot K \cdot m}$ where $C_{t,j}$ is the $j^{th}$ context vector in $\mathbb{R}^m$ at time step $t$  \\ 
$ A = \{arm_1,...,arm_N\}$  \\
$ NN_{1,Z_t}, NN_{2,Z_t} = $ Neural Network $:\mathbb{R}^m \rightarrow \mathbb{R} $ with $Z_t$ neurons \\
$ Reward : \mathbb{N}_{[1,K]} \rightarrow \mathbb{R}$ \\
 \For{t = 1 ... T}{
  \For{j = 1 ... K}{
   \# $NN(\cdot)$ gives a prediction using NN \\
   $ \hat{\mathbb{E}}_R(arm_j) = NN_{1,Z_t}(C_{t,j}) $ \quad \# Predict reward \\
   $ \hat{\mathbb{E}}_V(arm_j) = NN_{2,Z_t}(C_{t,j}) $ \quad \# Predict certainty \\
  }
  \For{j = 1 ... K}{
    \# Choose arms \\
    $D_{t,j} =  \arg \max_{arm \in A - D_{t,:} } \ \hat{\mathbb{E}}_R(arm)
    + \sqrt{\frac{\hat{\mathbb{E}}_V(arm) }{ t } } $ \\
  }
  \For{j = 1 ... K}{
    $R_{t,j} = Reward(D_{t,j})$ \quad \quad \quad \ \# Record rewards \\
    $V_{t,j} = (R_{t,j}-\hat{\mathbb{E}}_R(arm_j))^2 $ \quad \# Record accuracy \\
  }
  \# Train the networks \\
  TrainNNet($NN_{1,Z_t},\ in=C_{1:t,D_{1:t,1:K}},\ out=R_{1:t,1:K}, \ loss=MSE $)  \\
  TrainNNet($NN_{2,Z_t},\ in=C_{1:t,D_{1:t,1:K}},\ out=V_{1:t,1:K}, \ loss=L1 $) \\
 }
 \caption{Deep UCB2}
 \label{ducb2}
\end{algorithm}

\section{Experiments and Results}

We experiment on the top-$k$ ranking problem with two real world datasets. In the first dataset, MNIST \cite{mnist}, the algorithm is given $n$ images (corresponding to $n$ arms) and must choose the largest $k$ images to maximize the reward. The dataset is uniformly balanced so each image has equal chance to be any digit. The reward is the digit in the image. Normal noise is added to the reward to test how robust the algorithms are. In the second dataset, Mushrooms \cite{mushrooms}, the algorithm is given $n$ vectors representing mushrooms and must choose the most edible k mushrooms to maximize the reward. Each mushroom is either edible or not. The dataset is balanced so there are expected $k$ edible mushrooms in the $n$ mushrooms provided. The reward is 1 or 0 respectively. Normal noise is added to the reward to test robustness of the methods. We plot the cumlative reward normalized by time and the cumulative regret which is the sum of the regret at each time step. The regret is the expected maximum reward minus the reward awarded based on the choice of arms. We train all of the neural networks and models only every 20 iterations. The neural networks are trained over 20 epochs each time, have a single hidden layer of neurons, and have an initial learning rate that decreases 20\% each 4 epochs. To remove random sampling artifacts, we are plotting an average of 10 independent runs and the standard deviation across runs is consistent and approximately the same between methods.

\begin{figure}[h]
  \centering
  \includegraphics[scale=.75]{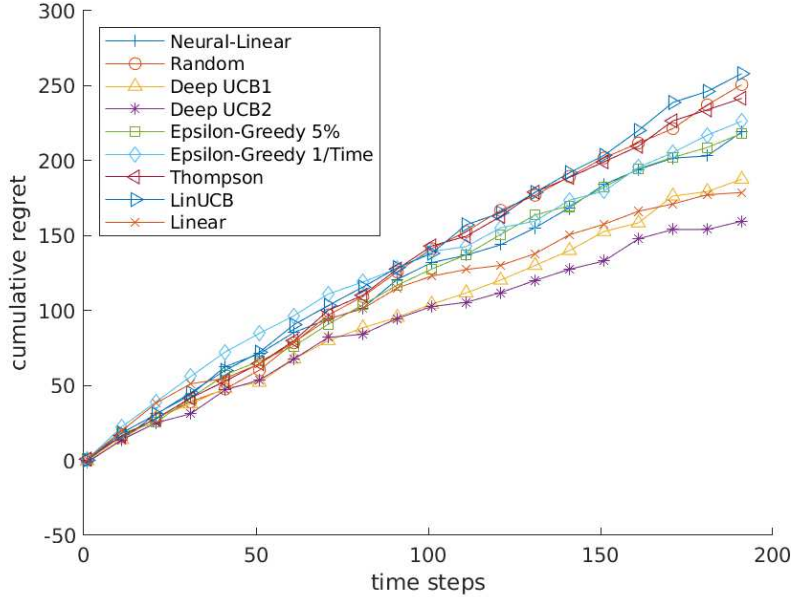}
  \caption{Top-$k$ ranking of Mushrooms. Arms=5. k=3. Reward noise = 2. Average of 10 runs plotting cumulative regret (lower is better). }
  \label{fig:mushroom_regret}
\end{figure}

In figure \ref{fig:mushroom_regret}, we plot the cumulative regret of each method on the Mushroom problem. For this problem, at each time step, a random mushroom context vector is supplied to each arm, which is edible/poisonous with equal probability. Then the method must choose the top-$k$ arms and receive a reward for each. The reward is 1 plus Gaussian noise for edible and 0 plus Gaussian noise for poisonous. The neural networks have a single hidden layer of 100 neurons, which is plenty here, and have initial learning rate 0.1. With no noise, this problem can easily be solved with a linear classifier. From the results, we see that with high noise, sigma=2, Deep UCB2 can outperform the linear regression method. We also note that the Deep UCB1/2 regret curves become flatter over time. This is what we would expect from the logarithmic bound, see Section 6. Indeed, the regret of all methods is always going up somewhat because of the random Gaussian noise in the reward that is impossible to predict. 

\begin{figure}[h]
  \centering
  \includegraphics[scale=.75]{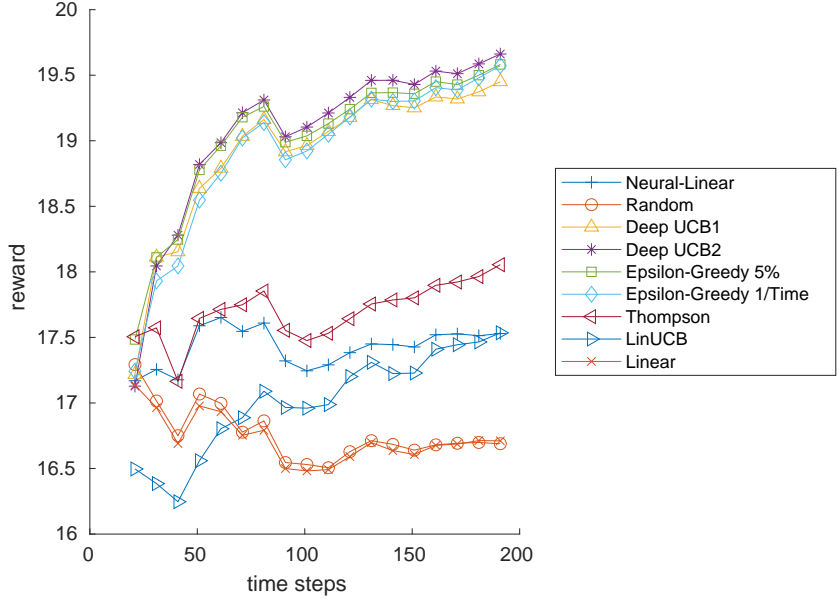}
  \caption{Top-$k$ ranking of MNIST images. Arms=5. k=3. Reward noise = 0.5. Average of 10 runs plotting cumulative reward normalized (divided by time step). }
  \label{fig:mnist_regret}
\end{figure}

In figure \ref{fig:mnist_regret}, we plot the cumulative reward normalized by time of each method for the MNIST problem. In the MNIST problem, at each time step, a random image is supplied to each arm, which is 0 to 9 with equal probability. Then the method must choose the top-$k$ arms and receive a reward for each. The reward is the image's digit plus Gaussian noise. The neural networks have a single hidden layer of 1000 neurons and have initial learning rate 0.001. We see that the methods Linear, LinUCB, and Neural-Linear underperform on this highly nonlinear problem. Thompson, with it's Gaussian prior assumption, also performs poorly since the distribution of the reward is highly non-Gaussian when the image is unknown. Epsilon-Greedy perform well but Deep UCB2 often outperforms since it explores more efficiently than Epsilon-Greedy's naive uniform exploration. 

\section{Regret Bounds Analysis}

Regret bounds show that a method will perform with limited regret, or limited mistakes, on an arbitrary problem or dataset. Bounds are ideal when employing a method in applications where many unknown problems be encountered. Here we will bound the regret of Deep UCB1. We specifically define the regret as the expectation of choosing the optimal arm minus the expectation of the method.  First, to bound the regret, we will need the neural networks to be able to learn the distribution/function. We use a classic theorem from \cite{gyorfi} which gives the convergence of the neural network in expectation with necessary technical assumptions.

\begin{theorem} 
\label{nnet_thm} [\cite{gyorfi} Theorem 16.3]
Let $NN_n$ be a neural network with $n$ parameters and the parameters are optimized to minimize MSE of the training data, $S = \{(X_i,Y_i)\}$ where $X$ is almost surely bounded. 
Let the training data, be size $n$, and $Y_i = R(x_i) \sim N(\mu_{x_i}, \sigma_{x_i})$ where 
$R : \R^m \rightarrow \R$. 
Then for $n$ large enough, 
$$ \E_S \int | NN_n(x) - \E(R(x)) |^2 dP(x) \le c\sqrt{\frac{\ln(n)}{n}} $$
for some $c > 0$.

\end{theorem}

This theorem implies that, with necessary assumptions, a trained neural network will learn the desired function with high probability. Next, we'll combine this with a UCB type proof. Auer et al. \cite{auer_finite-time_2002} showed the UCB1 algorithm achieves a logarithmic bound. We devise a similar proof for Deep UCB1 to show poly-logarithmic regret. Without loss of generality, we'll assume the $K-$armed bandit setting (that is top-1 MAB). Let $T_j(n)$, $1 \le j \le K$ be the random variable equal to number of times arm $j$ is chosen in the first $n$ steps. Define $I_n$ to be the arm chosen at time $n$.
Let $X$ be the context vector and $R(X)$ the stochastic reward. 
Denote $\mu_i(X) := \E(R_i(X))$ as the expectation of the reward $R_i(X)$ by choosing arm $i$ where $X$ is the given context vector. Further, define the notation $*$ for the optimal arm index. Thus $\mu_*(X)$ is the expectation of the optimal arm.

Define a \emph{Weak CMAB} to satisfy, for all $i : 1 \le i \le K, i \ne * $, that
$$ \mu_{*,min} - \mu_{i,max} - 2(\mu_{i,max} - \mu_{i,min}) = \delta_i > 0 $$
for some $\delta_i$, where $\mu_{i,max} := \sup_X \mu_{i}(X)$ and $\mu_{i,min} := \inf_X \mu_{i}(X)$. 
Let $\tilde \mu_i(X) = \E_X(NN_i(X))$ which is trained on $\{(X,R_i(X))\}$.

\begin{theorem} 
\label{deep_ucb_thm}
Assume that there is some $M_{NN} \ge \sup_X |NN(X)|$ for all possible neural nets (this is true with sigmoid activiation), the space $\mathcal{X} \subset \R^m$ containing $X$ is compact, $\mu_i$ continuous, and $W_t=\sqrt{t}$. With the assumptions of theorem \ref{nnet_thm}, Deep UCB1 method converges with square logarithmic regret,
$$ \sup_X \mu_{*}(X) n - \sum_{j=1}^K \mu_{j}(X) \E[T_j(n)] \le O(\ln^2(n)) $$
on Weak CMAB problems.
\end{theorem}

This theorem implies an optimal method in the sense that by dividing both sides by $n$, the lest hand side must converge to 0 with $n \rightarrow \infty$. So Deep UCB1 converges to the optimal solution in expectation with $n \rightarrow \infty$. Theorem proof in appendix. 

\section{Conclusion}
The CMAB problem, in general, has proved to be a difficult problem in recent decades. Using the new technology of neural networks and modern hardware, we can finally make progress on difficult datasets. We've introduced an improved method, Deep UCB, with mathematical guarantees and success on real world datasets. This new method is simple enough to be applied by practitioners in a variety of fields. 

\printbibliography

\section*{Appendix}

\begin{proof}[Proof of theorem \ref{deep_ucb_thm}]
Recall that $T_j(n)$, $1 \le j \le K$ is the random variable equal to number of times arm $j$ is chosen in the first $n$ steps. Define $I_n$ to be the arm chosen at time $n$.
Let $X$ be the context vector and $R(X)$ the stochastic reward.

Denote $\mu_i(X) := \E(R_i(X))$ as the expectation of the reward $R_i(X)$ by choosing arm $i$ where $X$ is the given context vector. Further, define the notation $*$ for the optimal arm index. For example, $\mu_*(X)$ is the expectation of the optimal arm. Define a \emph{Weak CMAB} to satisfy, for all $i : 1 \le i \le K, i \ne * $, that
$$ \mu_{*,min} - \mu_{i,max} - 2(\mu_{i,max} - \mu_{i,min}) = \delta_i > 0 $$
for some $\delta_i$, where $\mu_{i,max} := \sup_X \mu_{i}(X)$ and $\mu_{i,min} := \inf_X \mu_{i}(X)$. 
Let $\tilde \mu_i(X) = \E_X(NN_i(X))$ which is trained on $\{(X,R_i(X))\}$.
Let $\tilde \sigma_i(X)$ be a neural network estimate of the standard deviation of $R_i(X)$, $\sigma_i(R(X))$.
Let the average $\tilde NN_{i,k} = \frac{1}{\sqrt{k}}\sum_{i=1}^{\sqrt{k}} NN_i$ with each $NN$ trained on $k$ independent samples and $i$ is the associated arm index. Note that we have been dropping parameterization index of $NN$ for better readability. 

We assumed that there is some $M_{NN} \ge \sup_X |NN(X)|$ for all possible neural nets (this is true with sigmoid activiation). For each arm's reward $R_i$, set $J_{NN,i}$ be the threshold of $n$ in theorem \ref{nnet_thm} 
such that $ \E_S \int | NN_{i,n}(x) - \E(R(x)) |^2 dP(x) \le c\sqrt{\frac{\ln(n)}{n}} $
and furthermore $  | \E_S NN_{i,n}(x) - \E(R(x)) | \le c\sqrt{\frac{\ln(n)}{n}} \ \forall X $ (we assume $P(X)$ is positively supported). Let $J_{NN} = \max_i J_{NN,i}$.
Let 
$J = \max\{J_{NN}, \min \{q \in \mathbb{N} : c\sqrt{\frac{\ln(q)}{q}}\le \min_{i\ne *}\delta_{i}/4 \}\}$. 
Let $c_{t,i,s}(X) = \sqrt{(2|\tilde\sigma_i(X)|^2 + 2\ln t) / \sqrt{s}}$ and 
$\tilde c_{t,i,s}(X) = c_{t,i,s}(X) + \mu_{i,max} - \mu_{i,min}+2c\sqrt{\frac{\ln(J)}{J}}$ where $t$ is time, $i$ is arm index, and $s$ is number of samples collected. 
Then for any positive integer $l$ and $i \ne *$,
$$ T_i(n) = J + 1 + \sum_{t=K+1}^n \{I_t=i\} $$
$$ \le J +l + \sum_{t=K+1}^n \{I_t=i, T_i(t-1)\ge l \} $$
$$ \le J +l + \sum_{t=K+1}^n \{\tilde NN_{*,T_*(t-1)}(X_t) + \tilde c_{t-1,*,T_*(t-1)}(X_t) \le \tilde NN_{i,T_i(t-1)}(X_t)+\tilde c_{t-1,i,T_i(t-1)}(X_t), T_i(t-1)\ge l \} $$
$$ \le J +l + \sum_{t=K+1}^n \{\min_{0<s<t} \tilde NN_{*,s}(X_t) + \tilde c_{t-1,*,s}(X_t) \le \max_{l \le s_i < t} \tilde NN_{i,s_i}(X_t)+\tilde c_{t-1,i,s_i}(X_t) \} $$
$$ \le J +l + \sum_{t=1}^n \sum_{s=1}^{t-1} \sum_{s_i=l}^{t-1} \{ \tilde NN_{*,s}(X_t) + \tilde c_{t,*,s}(X_t) \le \tilde NN_{i,s_i}(X_t)+\tilde c_{t,i,s_i}(X_t) \} $$

The condition $\{ \tilde NN_{*,s}(X_t) + \tilde c_{t,*,s}(X_t) \le \tilde NN_{i,s_i}(X_t)+\tilde c_{t,i,s_i}(X_t) \} $ implies \\
$ \tilde NN_{*,s}(X_t) \le \tilde \mu_*(X_t) -  \tilde c_{t,*,s}(X_t) $ 
or $ \tilde NN_{i,s_i}(X_t) \ge \tilde \mu_i(X_t) +  \tilde c_{t,i,s_i}(X_t) $ \\
or $ \tilde \mu_*(X_t) < \tilde \mu_i(X_t) + 2\tilde c_{t,i,s_i}(X_t). $
By the Chernoff-Hoeffding bound

$$ P(\tilde NN_{*,s}(X_t) \le \tilde \mu_*(X_t) - \tilde c_{t,*,s}(X_t)) $$
$$ \le P(\tilde NN_{*,s}(X_t) \le \tilde \mu_{*,max} - \tilde c_{t,*,s}(X_t)) $$
$$ \le P(\tilde NN_{*,s}(X_t) \le \tilde \mu_{*,max} - c_{t,*,s}(X_t) - \mu_{*,max} + \mu_{*,min}-2c\sqrt{\frac{\ln(J)}{J}}) $$
$$ \le P(\tilde NN_{*,s}(X_t) \le \mu_{*,min} - c_{t,*,s}(X_t) -c\sqrt{\frac{\ln(J)}{J}}) $$
$$ \le P(\tilde NN_{*,s}(X_t) \le \tilde\mu_{*,min} - c_{t,*,s}(X_t) ) $$
$$ \le P(\tilde NN_{*,s}(X_t) \le \tilde \mu_{*,min} -\sqrt{(2\ \ln (\exp|\tilde\sigma_*(X_t)|^2 t))/\sqrt{s}} ) $$
$$ \le P(\sum_{i=1}^{\sqrt{s}} NN_{*,s}(X_t) \le \sqrt{s}\tilde \mu_{*,min} - \sqrt{s(2\ \ln (\exp|\tilde\sigma_*(X_t)|^2 t)) / \sqrt{s}} ) $$
$$ \le e^{-2 (\sqrt{s(2\ \ln (\exp|\tilde\sigma_*(X_t)|^2 t)) / \sqrt{s}})^2 / (\sqrt{s})} $$
$$ \le e^{-4 \ln (\exp|\tilde\sigma_*(X_t)|^2 t)} = (\exp|\tilde\sigma_*(X_t)|^2 t)^{-4} \le e^{0}\ t^{-4} = t^{-4} $$
and
$$ P(\tilde NN_{i,s_i}(X_t) \ge \tilde \mu_i(X_t) + \tilde c_{t,i,s_i}(X_t)) $$
$$ \le P(\tilde NN_{i,s_i}(X_t) \ge \tilde \mu_{i,min} + \tilde c_{t,i,s_i}(X_t)) 
\le t^{-4} $$

For $s_i$ large enough,
$$ \tilde \mu_*(X_t) -\tilde \mu_i(X_t)-2 \tilde c_{t,s_i}(X_t) $$
$$ \ge \mu_*(X_t) - c\sqrt{\frac{\ln(J)}{J}} - \mu_i(X_t) - c\sqrt{\frac{\ln(J)}{J}} - 2 \tilde c_{t,s_i}(X_t) $$
$$ \ge \tilde \mu_{*,min} - c\sqrt{\frac{\ln(J)}{J}} - \tilde \mu_{i,max} - c\sqrt{\frac{\ln(J)}{J}} - 2 \tilde c_{t,s_i}(X_t) $$

$$ \ge \tilde \mu_{*,min} - \delta_i/2 - \tilde \mu_{i,max} - [2\sqrt{(2|\tilde\sigma(X_t)|^2 + 2\ln t) }] s_i^{-1/4} - 2  \mu_{i,max} + 2 \mu_{i,min} - 4c\sqrt{\frac{\ln(J)}{J}} $$
$$ \ge \tilde \mu_{*,min} - \delta_i/2 - \tilde \mu_{i,max} - [2\sqrt{(2|\tilde\sigma(X_t)|^2 + 2\ln t) }] s_i^{-1/4} - 2 \tilde \mu_{i,max} + 2 \tilde \mu_{i,min} $$
$$ \ge \tilde \mu_{*,min} - \delta_i/2 - \tilde \mu_{i,max} - [2\sqrt{(2 M_{NN}^2 + 2\ln t) }] s_i^{-1/4} - 2 \tilde \mu_{i,max} + 2 \tilde \mu_{i,min} $$
$$ \ge \delta_i/2 - [2\sqrt{(2 M_{NN}^2 + 2\ln t) }] s_i^{-1/4} $$
$$ > 0 $$
Specifically, 
$$ s_i > 2^8\frac{ (2M_{NN}^2 + 2\ln t)^2 }
{ \delta_i^4} $$
Then set $l = \lceil 2^8\frac{ (2M_{NN}^2 + 2\ln n)^2 }
{ \delta_i^4} \rceil$ and the event is ruled out. So
$$ \E[T_i(n)] \le J + \lceil 2^8\frac{ (2M_{NN}^2 + 2\ln n)^2 }
{ \delta_i^4} \rceil $$
$$ + \sum_{t=1}^\infty \sum_{s=1}^{t-1} \sum_{s_i=l}^{t-1} P( \tilde NN_{*,s}(X_t) \le \mu_*(X_t) - c_{t,*,s}(X_t))+P(\tilde NN_{i,s_i}(X_t) \ge \mu_i(X_t) + c_{t,i,s_i}(X_t))$$
$$ \le J + \lceil 2^8\frac{ (2M_{NN}^2 + 2\ln n)^2 }
{ \delta_i^4} \rceil + \sum_{t=1}^\infty \sum_{s=1}^{t-1} \sum_{s_i=l}^{t-1} 2 t^{-4} $$
$$ \le J + 2^8\frac{ (2M_{NN}^2 + 2\ln n)^2 }
{ \delta_i^4} + 1 + \pi^2/3 $$
$$ \le O(\ln^2 n) + O(\ln n) $$

\end{proof}

\end{document}